\documentclass[runningheads]{llncs}
\usepackage{graphicx}
\usepackage{cite}
\usepackage{booktabs}
\usepackage{hyperref}
\usepackage{multirow}
\usepackage{caption}
\usepackage{xcolor}
\usepackage{amsmath}
\begin{document}
\title{Finding Significant Features for Few-Shot Learning using Dimensionality Reduction 
\thanks{{Currently under review for the Mexican Conference on AI (MICAI 2021)}}}
\titlerunning{Finding Significant Features for Few-Shot Learning}
%

\author{{Mauricio Mendez-Ruiz\inst{1} \and
Ivan Garcia\inst{2} \and
Jorge Gonzalez-Zapata\inst{2} \and \\Gilberto Ochoa-Ruiz \inst{1} \and Andres Mendez-Vazquez \inst{2} }}
\authorrunning{{M. Mendez et al.}}
%
\institute{Tecnológico de Monterrey, School of Engineering and Sciences, Mexico \and
CINVESTAV Unidad Guadalajara, Mexico \\
\email{\{A00812794, gilberto.ochoa\}@tec.mx}}

%
\maketitle
\begin{abstract}
Few-shot learning is a relatively new technique that specializes in problems where we have little amounts of data. The goal of these methods is to classify categories that have not been seen before with just a handful of samples. Recent approaches, such as metric learning, adopt the meta-learning strategy in which we have episodic tasks conformed by support (training) data and query (test) data. Metric learning methods have demonstrated that simple models can achieve good performance by learning a similarity function to compare the support and the query data. However, the feature space learned by a given metric learning approach may not exploit the information given by a specific few-shot task. In this work, we explore the use of dimension reduction techniques as a way to find task-significant features helping to make better predictions. We measure the performance of the reduced features by assigning a score based on the intra-class and inter-class distance, and selecting a feature reduction method in which instances of different classes are far away and instances of the same class are close. This module helps to improve the accuracy performance by allowing the similarity function, given by the metric learning method, to have more discriminative features for the classification. Our method outperforms the metric learning baselines in the miniImageNet dataset by around 2\% in accuracy performance.

\end{abstract}

\section{Introduction}

In recent years, we have witnessed the great progress of successful deep learning models and architectures \cite{Ganatra2018DeepLearning}, and the application in real-world problems. For example, in cases in Computer Vision, Natural Language Processing (NLP), speech synthesis, strategic games, etc. The new performance levels achieved by such deep architectures, have revolutionized those fields.
However, despite such advances in deep learning, the standard supervised learning does not offers a satisfactory solution for learning from small datasets (Few shot problem). This is due to the overfitting problem that Deep Learning incurs when a small dataset is used reducing their generalization capabilities.
Furthermore, there are many problem domains, such as health and medical settings, where obtaining labeled data can be very difficult or the amount of work required to obtain the ground truth representations is very large, time consuming and costly. 

An interesting phenomena is how humans deal with data scarcity and are able to make generalizations with few samples. Thus, it is desirable to reproduce these abilities in our Artificial Intelligence systems. In the case of Machine Learning, Few-Shot learning (FSL) methods has been proposed \cite{siamesenetworks, maml, prototypicalNets2017, matchingNetworks2016, sung2017relationNet} to imitate this ability by classifying unseen data from a few new categories. There are two main FSL approaches: The first one is Meta-learning based methods \cite{maml, learningtolearn, learningtooptimize, l2lgradient}, where the basic idea is to learn from diverse tasks and datasets and adapt the learned algorithm to novel datasets. The second are Metric-learning based methods \cite{distanceMetricLearning, siamesenetworks}, where the objective is to learn a pairwise similarity metric such that the score is high for similar samples and dissimilar samples get a low score. Later on, these metric learning methods started to adopt the meta learning policy to learn across tasks \cite{prototypicalNets2017, matchingNetworks2016, sung2017relationNet}.

The main objective of these methods is to learn an effective embedding network in order to extract useful features of the task and discriminate on the classes which we are trying to predict. From this basic learning setting, many extensions have been proposed to improve the performance of metric learning methods. Some of these works focus on pre-training the embedding network \cite{ssl}, others introduce task attention modules \cite{chen2020multiscale, categoryTraversal, principalCharacteristics}, whereas other try to optimize the embeddings \cite{convexoptimization} and yet others try to use a variety of loss functions \cite{principalCharacteristics}.

In this work, we focus on finding task-significant features by applying different feature reduction techniques and assigning the reduced features a score based on the inter and intra class separability. We believe that finding those relevant features for each task is important, as we can better discriminate between classes and obtain a better inference.

The rest of this paper is organized as follows. In Section \ref{sec:related_work} we introduce the related work and explain the problem setting.In Section \ref{sec:model} we introduce our proposed model, with the ICNN module which helps us to choose the best dimensionality reduction technique.Then, in Section \ref{sec:experiments} we give details on how we implemented our model, the design choices taken based on experiments and the comparison with baselines and state-of-the-art models. Finally, in Section \ref{sec:conclusion} we summarize our work and discuss future directions.


\section{Materials and Methods}
\label{sec:related_work}

\begin{figure}
\includegraphics[width=\textwidth]{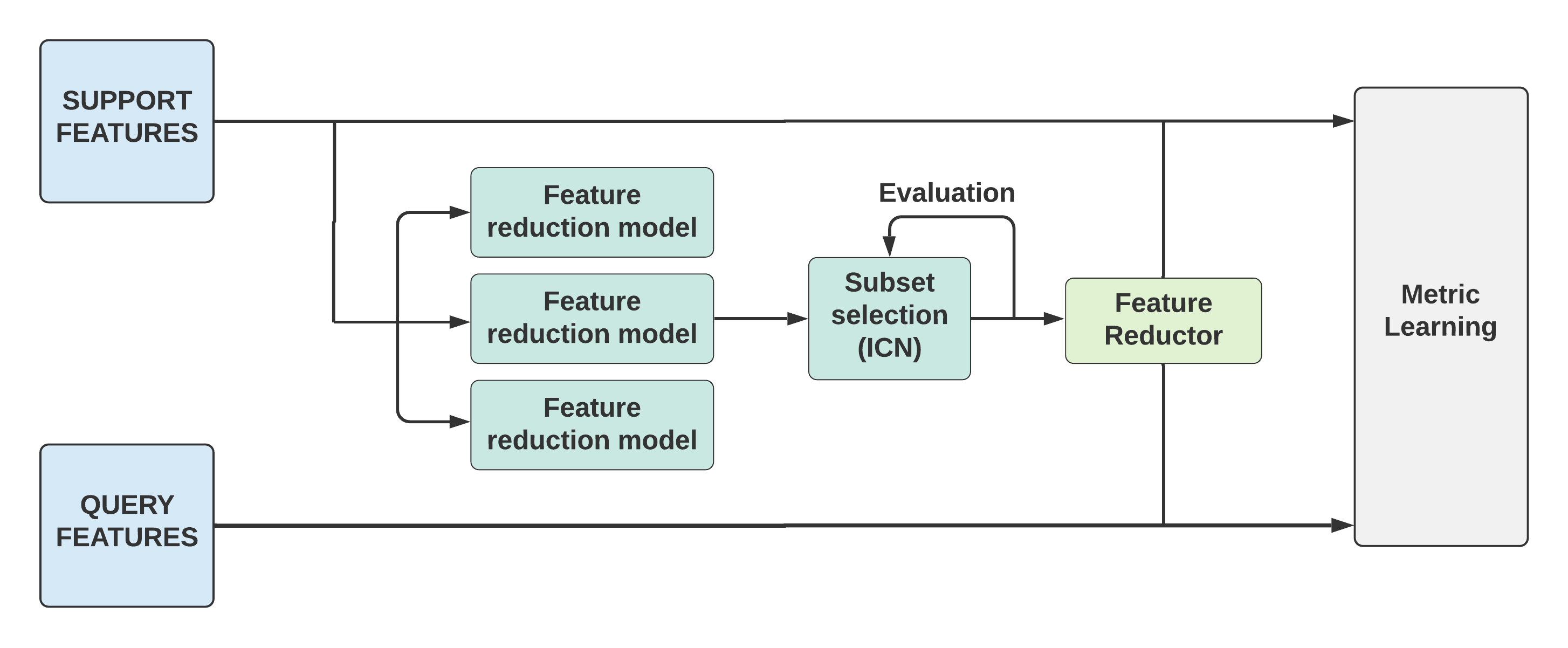}
\caption{The components of the (ICNNnet). After obtaining the features from the feature extractor, a number of feature reduction models are applied to the support embeddings. The features obtained from these models are measured via the ICNN Score to select the best feature reductor. Finally, this reductor is applied to the support and query embeddings to continue with the metric learning inference.} \label{fig:model}
\end{figure}

\subsection{Meta-learning tasks}

The few-shot meta-learning setup consists of episodic tasks, which can be seen as batches in traditional deep learning. A task is made up of support data and query data. The support set contains $k$ previously unseen classes and $n$ instances for each class, and the objective is to classify $q$ queries using the support data. This setting is also known as $k$-way $n$-shot (e.g. 5-way 1-shot or 5-way 5-shot). As described in \cite{matchingNetworks2016, optimizationfewshot}, the model is trained using an episodic mechanism, where each episode is loaded with a new random task taken from the training data. 

\subsection{miniImageNet dataset}

For the experimental results we used MiniImageNet \cite{matchingNetworks2016}, which is a subset of ImageNet version of ILSVRC-2012 \cite{imagenet} and is used as a benchmark for the evaluation of few-shot learning methods. This subset is comprised of 100 classes, each one containing 600 images, making up a total of 60,000 images. We follow the split proposed by Ravi and Larochelle \cite{optimizationfewshot}, dividing the dataset into 64 classes for training, 16 classes for validation and 20 classes for testing.

\section{Proposed Model}
\label{sec:model}

For our proposed model (see Figure \ref{fig:model}), we adopt a feature selection strategy based on dimensionality reduction assisted by an inter and intra class nearest neighbors distance score. As illustrated in Figure \ref{fig:model}, after obtaining the embeddings from the feature extractor we are left with a number of feature vectors, each one representing a task sample. From these features, we want to obtain the more relevant for the given task. We apply different feature reduction methods, and obtain an intra-class and inter-class score for each one. These scores are used to select the method which helps us to obtain the best dimensions for the current task. The obtained features are then used by a metric learner to produce a classification.

\subsection{Feature reduction techniques}\label{section:feature-models}

We selected the following feature reduction strategies to apply them with the feature vectors obtained from the few-shot learning task:

\begin{enumerate}
    \item Principal Component Analysis (PCA) \cite{pca}: 
    
    A dimensionality reduction method which transforms a large set of variables into a smaller one, while trying to preserve as much information of the data as possible.
    
    \item Uniform Manifold Approximation and Projection (UMAP) \cite{mcinnes2018umap}: 
    
    A manifold learning technique for non-linear dimension reduction that is mostly used for visualization purposes. This method tries to preserve the global structure of the data.
    
    \item Isometric mapping (Isomap) \cite{isomap}:
    
    A non-linear dimensionality reduction which seeks a low dimensional embedding which preserves the geodesic distances between the data points. The three stages of this algorithm are to create a neighborhood network, use a shortest path graph search to calculate the geodesic distance between all pair of points, and finally find the low dimensional embedding through eigenvalue decomposition.
    
\end{enumerate}


These methods cover the main attempts for feature generation from Linear Algebra, kernel methods and the use of random graphs for learning manifolds. Although other methods exist, as the t-SNE \cite{tsne}, they can be seen as particular applications of the previous methods. For example, the t-SNE defines a series of probability based on Gaussian kernels. Thus, using Kullback-Leiber a minimization procedure is executed to obtain a manifold based on those kernels. These types of minimization also happen in the PCA when using lagrange multipliers to obtain the eigenvectors and eigenvalues of the data covariance matrix. Later on section \ref{sec:experiments}, we show experiments made with more methods (Umap, PCA, Isomap, Kernel PCA, Truncated SVD, Feature Agglomeration, Fast ICA and Non-Negative Matrix Factorization) to prove that we do not need more methods than the three mentioned above.


\subsection{Inter and Intra Class Nearest Neighbors Score (ICNN Score)}

The ICNN Score \cite{tesis_ivan} was proposed to aid the feature selection based on supervised dimensionality reduction with subset evaluation. This measure improves the performance of dimensionality reduction techniques based on manifold algorithms by removing noisy features. There are two main concepts for the ICNN feature selection technique: Inter-class distance and Intra-class distance. Inter-class distance refers to the distance between points of different classes, and Intra-class distance refers to the distance between points of the same class. The idea for a successful feature selection approach is to choose those which increase the inter-class distance and reduce the intra-class distance, in order to allow the task to be differentiated.

The ICNN Score is a measure that combines the distance and variance of the inter-intra k-nearest neighbors of each instance in the data. The formula to calculate this score is the following:

\begin{equation}
    ICNN(X) = \frac{1}{\left | X \right |} \sum_{x_i \in X} \lambda(X_i)^\frac{1}{p} \omega(X_i)^\frac{1}{q} \gamma(X_i)^\frac{1}{r},
\end{equation}where $p$, $q$ and $r$ are control constants. 

Here, $\lambda$ is a function that penalizes the neighbors of $X_i$ with the same class based on how distant they are, and the neighbors of different classes based on how close they are:

\begin{equation}
    \lambda(X_i) = \frac{\sum_{p\in K_{\tilde{x}_i}} \frac{d(X_i, p) - \theta(X_i)}{\alpha(X_i) - \theta(X_i)} + \sum_{q\in K_{x_i}} 1 - \frac{d(X_i, q) - \theta(X_i)}{\alpha(X_i) - \theta(X_i)}}{\left | K_{x_i} \right | + \left | K_{\tilde{x}_i} \right |},
\end{equation} where $K_{x_i} = KNN(x_i) \in y_i$ are the set of k-nearest neighbors of $x_i$ that have the same class.
$K_{\tilde{x}_i} = KNN(x_i) \in y_j \neq y_i$ are the set of k-nearest neighbors of $x_i$ that has different class.
$d(a, b)$ is a distance function, which in this case is the euclidean distance.
$\alpha(X_i)$ and $\theta(X_i)$ are the maximum distance and the minimum distance of the $x_i$ neighbors, respectively.

In the ideal scenario, the neighbor's distance of the same class are close to 0 and the distance with different classes are close to 1. 

Now, $\omega$ is a function that penalizes the distance variance of neighbors (Eq. \ref{omega}):

\begin{align}
\label{omega}
\begin{split}
    \omega(X_i) = 1 - & \bigg(Var\Big(\sum_{p\in K_{\tilde{x}_i}} \frac{d(X_i, p) - \theta(X_i)}{\alpha(X_i) - \theta(X_i)}\Big) + \\ 
    & Var\Big(\sum_{q\in K_{x_i}} 1 - \frac{d(X_i, q) - \theta(X_i)}{\alpha(X_i) - \theta(X_i)}\Big)\bigg),
\end{split}
\end{align} 
where a high variance is penalized because it increases the possibility of overlapping classes.

\begin{figure}
\minipage{0.32\textwidth}
  \includegraphics[width=\linewidth]{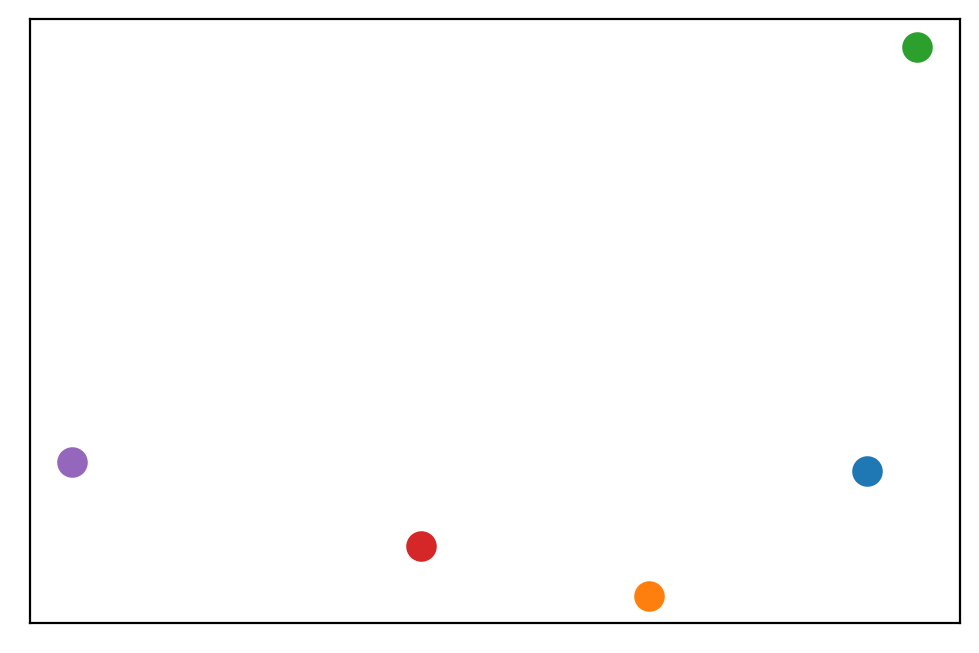}
  \caption*{ICNN Score: 0.71}
\endminipage\hfill
\minipage{0.32\textwidth}
  \includegraphics[width=\linewidth]{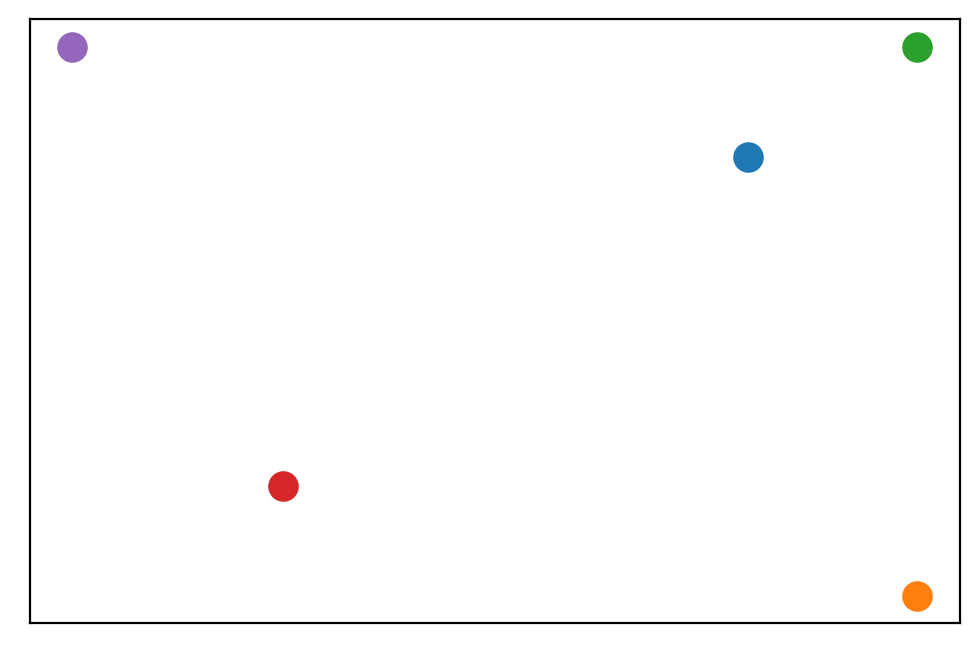}
  \caption*{ICNN Score: 0.88}
\endminipage\hfill
\minipage{0.32\textwidth}
  \includegraphics[width=\linewidth]{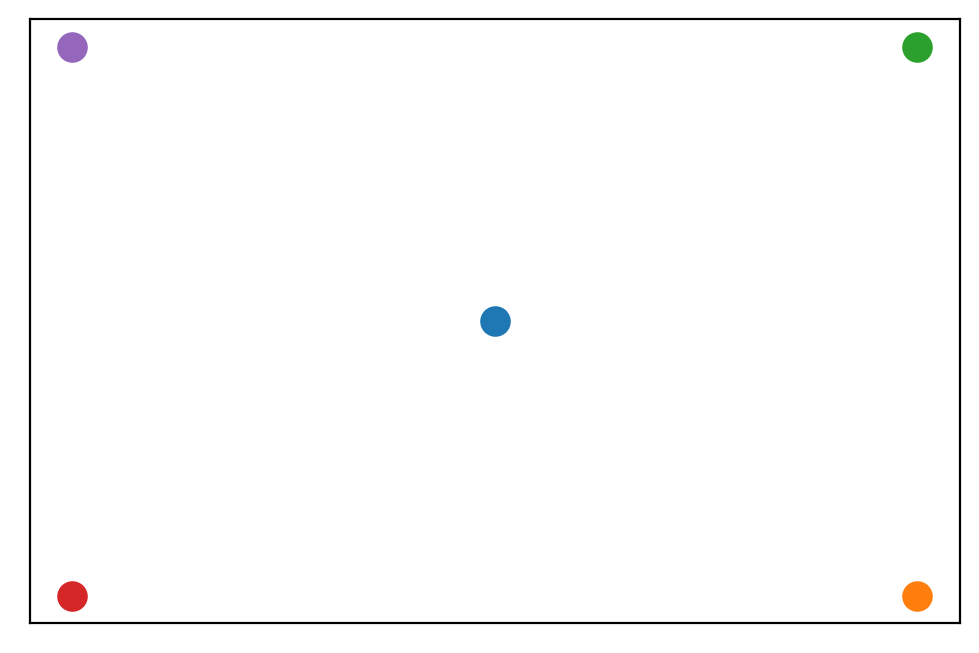}
  \caption*{ICNN Score: 0.97}
\endminipage
\caption{ICNN score behaviour on different 5-way 1-shot scenarios. The hyper-parameters for calculating the score are set to: $k=3$, $p=2$, $q=2$, $r=2$}
\label{fig:5way1shot_icnn}
\end{figure}

\begin{figure}[]
\minipage{0.32\textwidth}
  \includegraphics[width=\linewidth]{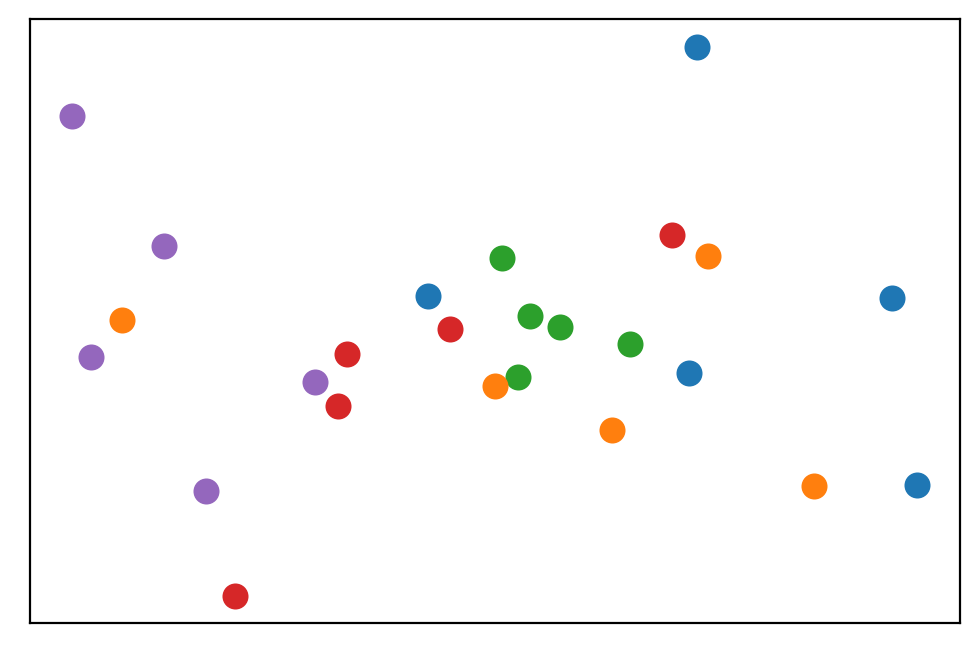}
  \caption*{ICNN Score: 0.47}
\endminipage\hfill
\minipage{0.32\textwidth}
  \includegraphics[width=\linewidth]{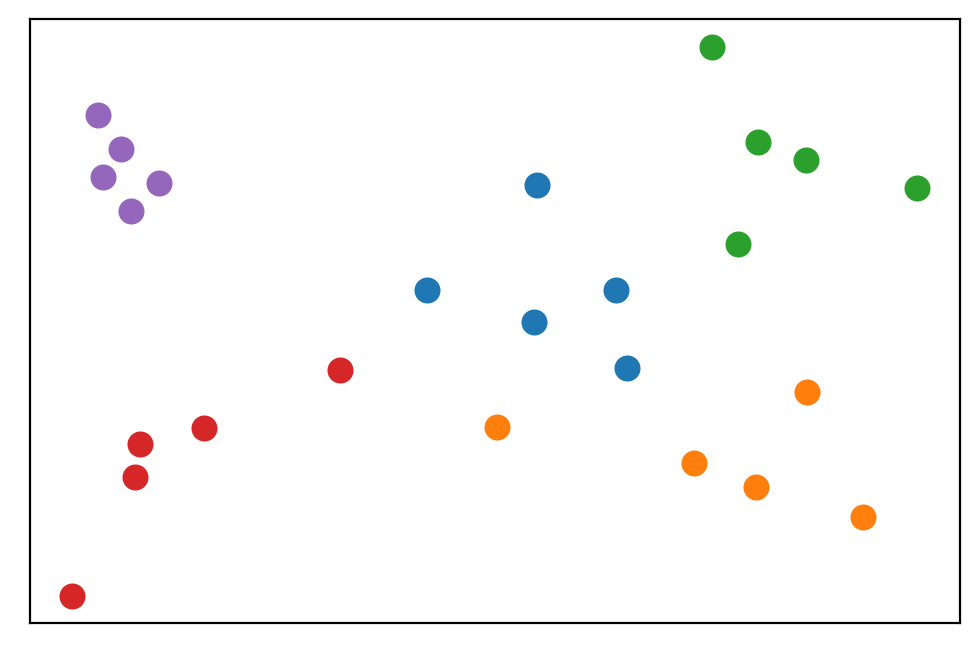}
  \caption*{ICNN Score: 0.72}
\endminipage\hfill
\minipage{0.32\textwidth}
  \includegraphics[width=\linewidth]{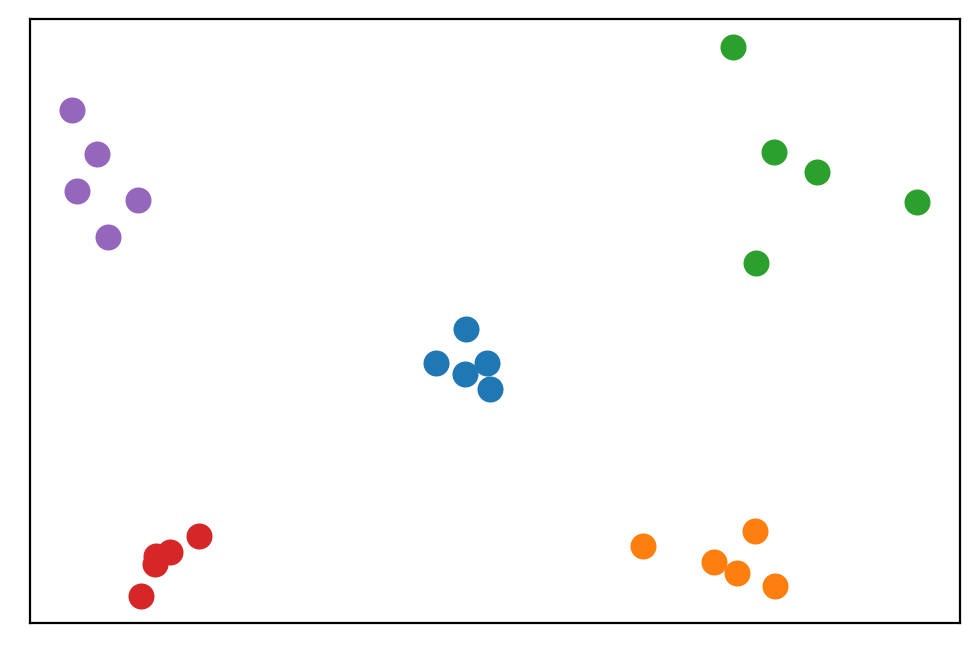}
  \caption*{ICNN Score: 0.96}
\endminipage
\caption{ICNN score behaviour on different 5-way 5-shot scenarios. The hyper-parameters for calculating the score are set to: $k=5$, $p=2$, $q=2$, $r=2$}
\label{fig:5way5shot_icnn}
\end{figure}

Finally, the $\gamma$ function describes the ratio of the neighbor's classes:

\begin{equation}
    \gamma(x_i) = \frac{\left | K_{x_i} \right |}{\left | K_{x_i} \right | + \left | K_{\tilde{x}_i} \right |},
\end{equation}
where each instance is penalized based on the neighbors in the same class of $x_i$. Each of the three functions ($\lambda, \omega$ and  $\gamma$) have an output with a range between 0 and 1. In figure \ref{fig:5way1shot_icnn} and \ref{fig:5way5shot_icnn}, we can visualize how some few-shot tasks scenarios would be rated using the ICNN score.

Using this metric, we can evaluate each feature reduction technique, as well as the original feature vector, to choose the best selection of features that are relevant for the current task.

\section{Experiments}
\label{sec:experiments}


\textbf{Evaluation Metric.} Following most of the metric learning methods \cite{prototypicalNets2017, matchingNetworks2016, sung2017relationNet}, we report our results on the mean accuracy (\%) over 1,000 test episodes with 95\% of confidence intervals.

\subsection{Implementation Details.} 

\textbf{Feature Extractor.} For the feature extractor, we test with two different backbones: ConvNet and ResNet-12. The ConvNet architecture follows the setting used by Vinyals et al. \cite{matchingNetworks2016}, with 4 layers of convolutional blocks. Each block is composed of a 3 $\times$ 3 convolution with 64 filters, followed by Batch Normalization and a ReLU layer. This network is optimized with Adam optimizer with an initial learning rate of $10^{-3}$. For the ResNet-12, following recent work \cite{categoryTraversal, convexoptimization}, the network is pre-trained using the SGD optimizer with momentum of 0.9 and learning rate of 0.1 over 100 epochs with a batch size of 128. Then, we apply the meta-training using SGD optimizer with momentum of 0.9 and learning rate of 0.0001.

\textbf{Meta-learning setup.} In order to compare against the baselines, our experiments are made under the 5-way 1-shot and 5-way 5-shot setting with 15 query images for each class in the task.  All the input images are resized to 84 $\times$ 84. On the training phase, we randomly construct 100 tasks over 200 epochs and apply validation over 500 tasks after every epoch. We train the network and obtain the cross-entropy loss. The initial learning rate is reduced by half every 20 epochs. For the testing phase, we randomly construct 1,000 tasks and measure the mean accuracy with 95\% confidence intervals.

\subsection{Model design choices}
\label{subsec_design_choices}

\begin{table}[t]
\centering
\setlength{\tabcolsep}{6pt}
\renewcommand{\arraystretch}{1.5}
\begin{tabular}{@{}lccc@{}}
\toprule
\multicolumn{1}{c}{\textbf{(\# of components; Type; Model)}} & \textbf{BackBone} & \textbf{1shot} & \textbf{5shot} \\ \midrule
Prototypical Networks (Paper)               & ConvNet   & 49.52          & 68.20          \\
(1) 6 Components; Support\&Query; Base        & ConvNet   & 52.37          & \textbf{69.08} \\
(2) Multiple Components; Support\&Query; Base & ConvNet   & \textbf{52.43} & 68.77          \\
(3) 6 Components; Support; Base          & ConvNet   & 46.88          & 38.24          \\
(4) 6 Components; Support\&Query; All         & ConvNet   & 51.44          & 67.26          \\
(5) Multiple Components; Support\&Query; All  & ConvNet   & 51.34          & 67.08          \\ \midrule
Prototypical Networks (our implementation)  & ResNet-12 & 61.13          & 76.21          \\
(1) 6 Components; Support\&Query; Base        & ResNet-12 & 63.03          & \textbf{78.14} \\
(2) Multiple Components; Support\&Query; Base & ResNet-12 & 62.30          & 78.12          \\
(3) 6 Components; Support; Base          & ResNet-12 & 56.67          & 77.37          \\
(4) 6 Components; Support\&Query; All         & ResNet-12 & \textbf{63.81} & 76.68          \\
(5) Multiple Components; Support\&Query; All  & ResNet-12 & 63.19          & 76.93          \\ \bottomrule
\end{tabular}
\caption{Test accuracies of the design choices experiments for 5-way tasks. See section \ref{subsec_design_choices} for more details}
\label{tab:design-choices}
\end{table}

Having the setup for the model extension stated above, we now discuss the ICNN hyper-parameters and feature reduction techniques settings that we chose to test in order to find the best combination that will give us better accuracy performance. 

The first set of of ablation studies is carried out in order to find the best hyper-parameters for the ICNN score. There are four constants that we need to choose for the algorithm, $k$, $p$, $q$ and $r$. We decided to give the same weight for $\lambda, \omega$ and $\gamma$. For this, we take the decision to use the same value for those three constants and set its value to $2$. For the $k$ (k - Nearest Neighbors), we decide to assign it a value related to the few-shot task. Since in the given task we have $c$ classes and $n$ shots, we decide to set to $k$ the value of $n$. In this way, we can ensure that, for each point, it always appears a nearest neighbor with different class. By having always a nearest neighbor of different class, we can obtain a better estimation in the $\lambda$ function, since we have now a perspective of the neighbors in the same class in relation with those of other classes.  For the case of the 1-shot setting, we decide to set $k$ to $3$, since all the other data points are of different class and this allows us to better understand the inter-class distance.


For the second set of ablation studies for setting the design choices, we study the effect of the variations in the feature dimensionality reduction techniques. There are three main concepts that we test related to the dimension reduction strategies:

\begin{figure}[t]
\includegraphics[width=\textwidth]{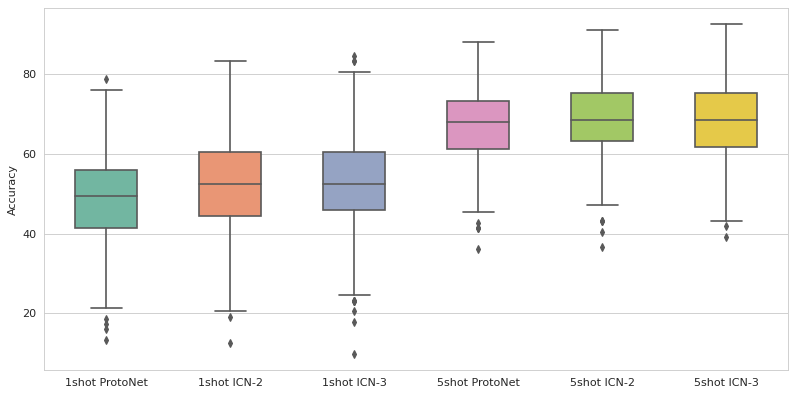}
\caption{Accuracy results over 1000 test episodes for the Prototypical Networks and our best proposed models using a ConvNet as a Feature Extractor, on the 1-shot and 5-shot setting. The ICN model number corresponds to the design choice showed on table \ref{tab:design-choices}.} \label{fig:convnet_boxplot}
\end{figure}

\begin{enumerate}
    \item The number of components to which we reduce the feature embedding vectors. In this case, we started by testing the reduction to 4, 6, 8 and 10 components. We found out that it was giving almost the same results, but it was a little better with 6 components. For the next experiments, we tested reducing to multiple components by halfs depending on the feature extractor. For the ConvNet, the feature embedding has 64 dimensions, and the reduced components were 32, 16, 8 and 4. For the ResNet-12, the feature embedding has 512 dimensions, and the reduced components were 256, 128, 64, 32, 16 and 8.
    \item The set of points used for the reduction (support / support \& query). We added these experiments to tests if only the support data was being useful for obtaining a good reduction, or we could aid the reduction by adding the query data.
    \item The set of feature reduction techniques used. Here we tested two different settings: using only PCA, Isomap and Umap (Base), or using all the models stated in section \ref{section:feature-models} (All). The base setting was obtained by testing the model on 1000 episodes and keeping the three feature reduction techniques that were chosen the most.
\end{enumerate}

There are some findings obtained from these experiments. We found that using UMAP in our feature reduction models, the training phase execution time greatly increased. For this reason, we decided to remove UMAP from the methods used in training, and use it only on the testing phase. 

\begin{figure}
\includegraphics[width=\textwidth]{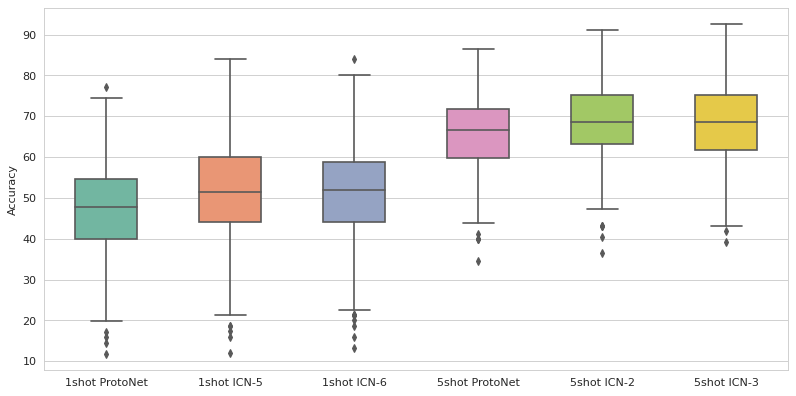}
\caption{Accuracy results over 1000 test episodes for the Prototypical Networks and our best proposed models using a ResNet-12 as a Feature Extractor, on the 1-shot and 5-shot setting. The ICN model number corresponds to the design choice showed on table \ref{tab:design-choices}.} \label{fig:resnet_boxplot}
\end{figure}

By testing the number of components, we first found that 6 components gave us slightly better results than reducing to 4, 8 and 10 components. Then, we also tested the model by reducing dimensions to multiple components. We found that, in most cases, reducing only to 6 components was giving better results. We believe that 6 components are good to keep most of the main information from the task while giving a better representation of the data.

With the experiments on the set used for reducing the dimensions, we found that using only the support data give us bad results. This proves that, on the few-shot settings tested (5-way 1-shot and 5-way 5-shot), the support data is not enough to obtain a good representation of the data after reducing the dimensions with the different feature reduction techniques. Using the support and query data allowed the feature reduction methods to better interpret the structure of the data, thus obtaining a better ICNN score.

As for the feature reduction techniques, we found that using the base models (PCA, Isomap and Umap) give us better results than using all the reduction models, on most of the experiments. This could be happening because using the worst behaving models may be resulting in wrong decisions for the prototypes. Further work can focus on visualizing the features obtained by each feature reduction technique.

The results of these experiments are summarized in table \ref{tab:design-choices}. We also provide a visualization for the behaviour of the best models obtained with the experiments stated above. In figure \ref{fig:convnet_boxplot}, we can visualize the accuracies from 1000 episodes obtained using the original prototypical networks and the best models obtained adding the ICNN module, all of these using the ConvNet as feature extractor and with the 5-way 1-shot and 5-way 5-shot setting. We can see improvements on the first and fourth quartile, and in some cases we have a smaller box from the second and third quartile, which means that the classification become a little more robust. 

The same visualization is showed in figure \ref{fig:resnet_boxplot}, but these accuracies are obtained using a ResNet-12 as feature extractor. We can see here that the first and second quartile are improving, with a better improvement on the first quartile. We can also see that, while the 1-shot setting gets a better improvement on the mean accuracy, the 5-shot setting obtain a better improvement on the first quartile.

\begin{table}[t]
\centering
\setlength{\tabcolsep}{10pt}
\renewcommand{\arraystretch}{1.5}
\begin{tabular}{@{}r|ccc@{}}
\toprule
\textbf{Model}                & \textbf{Network} & \textbf{1-shot}           & \textbf{5-shot}           \\ \midrule
Matching Networks             & ConvNet                    & 43.56 $\pm$ 0.84          & 55.31 $\pm$ 0.73          \\
Prototypical Networks         & ConvNet                    & 49.42 $\pm$ 0.78          & 68.20 $\pm$ 0.66          \\
Relation Networks             & ConvNet                    & 50.44 $\pm$ 0.82          & 65.32 $\pm$ 0.70          \\ \midrule
ProtoNets + ICNN & ConvNet & \multicolumn{1}{l}{\textbf{52.43 $\pm$ 0.73}} & \multicolumn{1}{l}{\textbf{69.08 $\pm$ 0.76}} \\ \midrule
K-tuplet Nets                 & ResNet-12                  & 58.30 $\pm$ 0.84          & 72.37 $\pm$ 0.63          \\
ProtoNets + CTM               & ResNet-12                  & 59.34 $\pm$ 0.55          & 77.95 $\pm$ 0.06          \\
Principal Characteristic Nets & ResNet-12                  & 63.29 $\pm$ 0.76          & 77.08 $\pm$ 0.68          \\ \midrule
ProtoNets + ICNN              & ResNet-12                  & \textbf{63.81 $\pm$ 0.71} & \textbf{78.14 $\pm$ 0.50} \\ \bottomrule
\end{tabular}
\caption{Comparison of test accuracies for 5-way tasks. See section \ref{comparison_baselines} for more details.}
\label{tab:comparison-stateoftheart}
\end{table}

\subsection{Comparison with baselines}
\label{comparison_baselines}

To validate the effectiveness of our new module, we compare it against similar state-of-the-art models following the standard few-shot learning setting. First, we compare it against the three main metric learning methods:  (1) Prototypical Networks \cite{prototypicalNets2017}, (2) Matching Networks \cite{matchingNetworks2016} and (3) Relation Networks \cite{sung2017relationNet}. These three models use a ConvNet as feature extractor, which obtains a feature embedding of size 64. We also compare against Category Traversal Module (CTM) \cite{categoryTraversal}, a model with the same idea of looking for task-relevant features, Principal Characteristic Network \cite{principalCharacteristics} and K-Tuplets Network \cite{ktuplets}. These three models use a ResNet-12 as feature extractor, with 512 dimensions in the output feature embedding.

Table \ref{tab:comparison-stateoftheart} illustrate the comparison of all the previous mentioned models with our method. We obtained an improvement of around 2\% for the 5-way 1-shot setting on the test set using Prototypical Networks. We also achieved a better performance than Matching Nets and Relation Nets on 5-way 5-shot setting by a large margin, but obtained a little improvement of around 1\% compared with Prototypical Nets. As for the other three models with ResNet-12 as feature extractor, we obtained an improvement of around 1\% on both settings of 1-shot and 5-shot.

\section{Conclusion}
\label{sec:conclusion}

In this paper, we propose a new module with the purpose of finding task-significant features by using dimensionality reduction techniques and a discriminant score based on intra and inter class nearest neighbors. The performance of the proposed model improves the accuracy compared to the metric learning baselines. We also compare the results with state-of-the-art models with deeper backbone and obtain a gain in accuracy performance. 

Our experiments are based on the combination of Prototypical Networks and ICNN but, as this method is proposed to obtain better features, any other metric learning technique (Matching Networks, Prototypical Networks) is expected to improve. The experimentation of our ICNN module with these other techniques are left for future work. 



\bibliographystyle{splncs04}
\bibliography{references}

%
%
%
%

\end{document}